\pdfoutput=1

\PassOptionsToPackage{table}{xcolor}

\documentclass[11pt]{article}

\usepackage[preprint]{acl}

\usepackage[T1]{fontenc}
\usepackage[utf8]{inputenc}
\usepackage{times}
\usepackage{latexsym}
\usepackage{inconsolata}
\usepackage{microtype}

\usepackage{booktabs}
\usepackage{array}
\usepackage{multirow}
\usepackage{enumitem}
\usepackage{authblk}
\usepackage{tcolorbox}

\usepackage{amsmath}
\usepackage{graphicx}
\usepackage{colortbl}

\definecolor{verylowscore}{RGB}{235,245,255}
\definecolor{lowscore}{RGB}{198,219,239}
\definecolor{midscore}{RGB}{107,174,214}
\definecolor{highscore}{RGB}{33,113,181}

\definecolor{lowcaq}{RGB}{208,209,230}
\definecolor{midcaq}{RGB}{128,125,186}
\definecolor{highcaq}{RGB}{84,39,143}
\definecolor{royalblue}{rgb}{0.5647, 0.2118, 0.6118}
\definecolor{lightpink}{rgb}{1.0, 0.71, 0.76}

\newcommand{\rot}[1]{\multicolumn{1}{c}{\rotatebox{90}{#1}}}

\newcommand{\scorecell}[1]{%
  \ifnum\pdfstrcmp{#1}{0.8}<0
    \cellcolor{verylowscore}#1%
  \else\ifnum\pdfstrcmp{#1}{0.7}<0
    \cellcolor{lowscore}#1%
  \else\ifnum\pdfstrcmp{#1}{0.6}<0
    \cellcolor{midscore}#1%
  \else
    \cellcolor{highscore}\textcolor{white}{#1}%
  \fi\fi\fi
}

\newcommand{\caqcell}[1]{%
  \ifnum\pdfstrcmp{#1}{0.7}<0
    \cellcolor{lowcaq}\textbf{#1}%
  \else\ifnum\pdfstrcmp{#1}{0.8}<0
    \cellcolor{midcaq}\textbf{\textcolor{white}{#1}}%
  \else
    \cellcolor{highcaq}\textbf{\textcolor{white}{#1}}%
  \fi\fi
}

\newcommand{\equalcontrib}{\textsuperscript{*}}

\title{CliME: Evaluating Multimodal Climate Discourse on Social Media and the \textit{Climate Alignment Quotient (CAQ)}}

\makeatletter
\def\@fnsymbol#1{\ensuremath{\ifcase#1\or *\or \dagger\or \ddagger\or
   \mathsection\or \mathparagraph\or \|\or **\or \dagger\dagger
   \or \ddagger\ddagger \else\@ctrerr\fi}}
\makeatother

\author{
  Abhilekh Borah\equalcontrib$^1$,
  Hasnat Md Abdullah\equalcontrib$^2$,
  Kangda Wei$^2$,
  Ruihong Huang$^2$ \\
  $^1$Manipal University Jaipur, India \\
  $^2$Texas A\&M University, USA \\ \\
  \texttt{abhilekh.229301149@muj.manipal.edu, hasnat.md.abdullah@tamu.edu,} \\
  \texttt{kangda@tamu.edu, huangrh@cse.tamu.edu} \\
}

\begin{document}
\maketitle

\renewcommand{\thefootnote}{\fnsymbol{footnote}}
\footnotetext[1]{These authors contributed equally to this work.}
\begin{abstract}
The rise of Large Language Models (LLMs) has raised questions about their ability to understand climate-related contexts. Though climate change dominates social media, analyzing its multimodal expressions is understudied, and current tools have failed to determine whether LLMs amplify credible solutions or spread unsubstantiated claims. To address this, we introduce \textbf{CliME (Climate Change Multimodal Evaluation)}, a first-of-its-kind multimodal dataset, comprising 2579 Twitter and Reddit posts. The benchmark features a diverse collection of humorous memes and skeptical posts, capturing how these formats distill complex issues into viral narratives that shape public opinion and policy discussions. To systematically evaluate LLM performance, we present the \textbf{Climate Alignment Quotient (CAQ)}, a novel metric comprising five distinct dimensions: \textbf{Articulation,  Evidence, Resonance, Transition,} and \textbf{Specificity}. Additionally, we propose three analytical lenses: \textbf{Actionability, Criticality,} and \textbf{Justice}, to guide the assessment of LLM-generated climate discourse using CAQ. Our findings, based on the CAQ metric, indicate that while most evaluated LLMs perform relatively well in Criticality and Justice, they consistently underperform on the Actionability axis. Among the models evaluated, Claude 3.7 Sonnet achieves the highest overall performance. We publicly release our \href{https://huggingface.co/datasets/climedataset/CliME}{\textcolor{royalblue}{\textit{CliME}}} dataset and \href{https://github.com/abhilekhborah/CliME}{\textcolor{royalblue}{\textit{code}}} to foster further research in this domain.
\end{abstract}


\begin{figure}[t]
    \centering
    \includegraphics[scale=0.5, trim={0.2cm 1.5cm 0cm 0cm}]{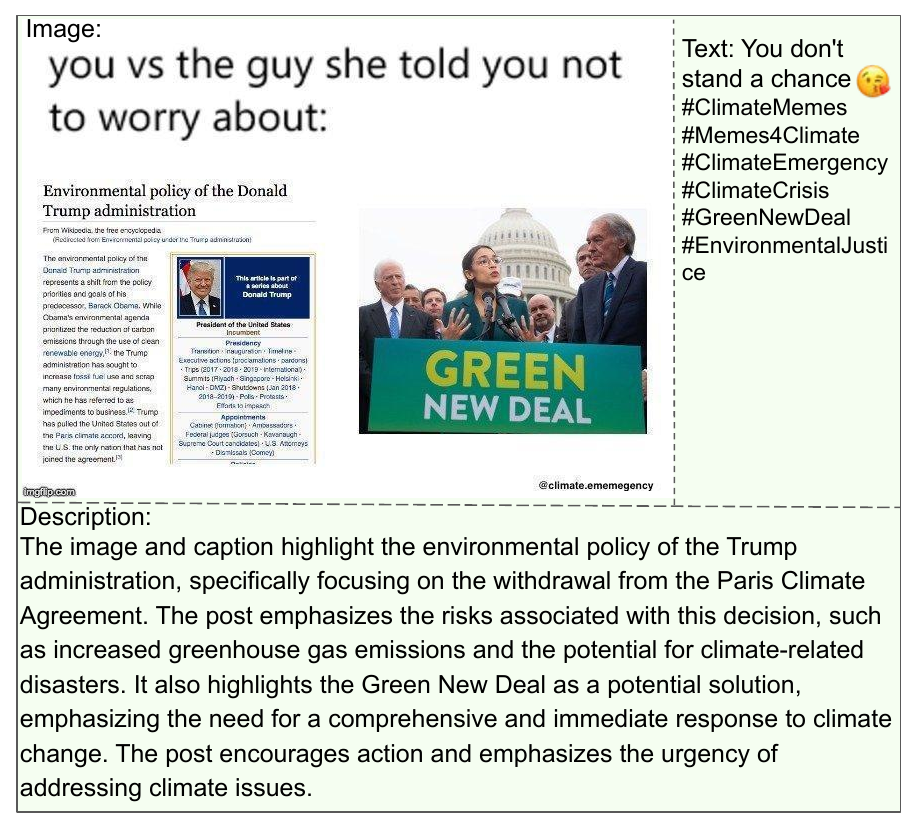}
    \vspace{-0.1cm}
    \caption{CliME sample data: Each data point includes a climate change-related image from a Reddit or Twitter post, the accompanying post text, and a generated description integrating both the image and text.}
    \label{fig:main}
    \vspace{-0.5cm}
\end{figure}

\section{Introduction}
Climate change has emerged as one of the most pressing challenges of our time, not only in scientific and policy circles but also in the public imagination \cite{change2018global, baste2021making, yusoff2011climate}. A 2023 study found that YouTube had the most significant positive effect on climate awareness in Latin America, followed by Instagram and Twitter, demonstrating the role of these platforms in disseminating climate-related information \cite{gomez2023effect}.
Social media platforms like Twitter (now X) and Reddit have emerged as prime spaces for climate discourse, shaping public opinion, mobilizing activism, and influencing policy. Viral campaigns such as \textbf{\#FridaysForFuture}, \textbf{\#ShowYourStripes} and \textbf{\#EarthHour} illustrate the power of social networks in transforming grassroots efforts into global movements. However, these online platforms propagate misinformation and polarization, vastly associated with skepticism, contrarianism, and denial \cite{treen2020online}. In 2023 alone, posts denying climate change on X tripled compared to previous years, highlighting the persistence of climate denial narratives online \cite{climateusatoday}. The rise of multimodal content in this domain further complicates this landscape. A flooded city image paired with \textit{“Climate policies harm the economy” }spreads faster than peer-reviewed data, exploiting visual-emotional resonance \cite{o2020more}. The rising hype of LLMs show their outstanding capabilities at text generation but their proficiency in grounding responses in visual-textual context in climate communication, remains unproven. Early studies reveal concerning trends: when prompted with climate-skeptical posts, models like GPT-4 often default to vague, non-committal language. For instance, a study by \textit{NewsGuard} found that GPT-4 \cite{hurst2024gpt} was more likely than its predecessor GPT-3.5 \cite{gpt35} to generate misinformation when prompted with false narratives, including those about climate change, without sufficient disclaimers or corrections \cite{axios, desmog}. This mirrors the "false balance" tactic often used to undermine scientific consensus, where both sides of an issue are presented as equally valid despite overwhelming evidence favoring one side. 

\begin{figure*}[t]
    \centering
    \includegraphics[scale=0.66, trim={0.25cm 0.5cm 0cm 0.5cm}]{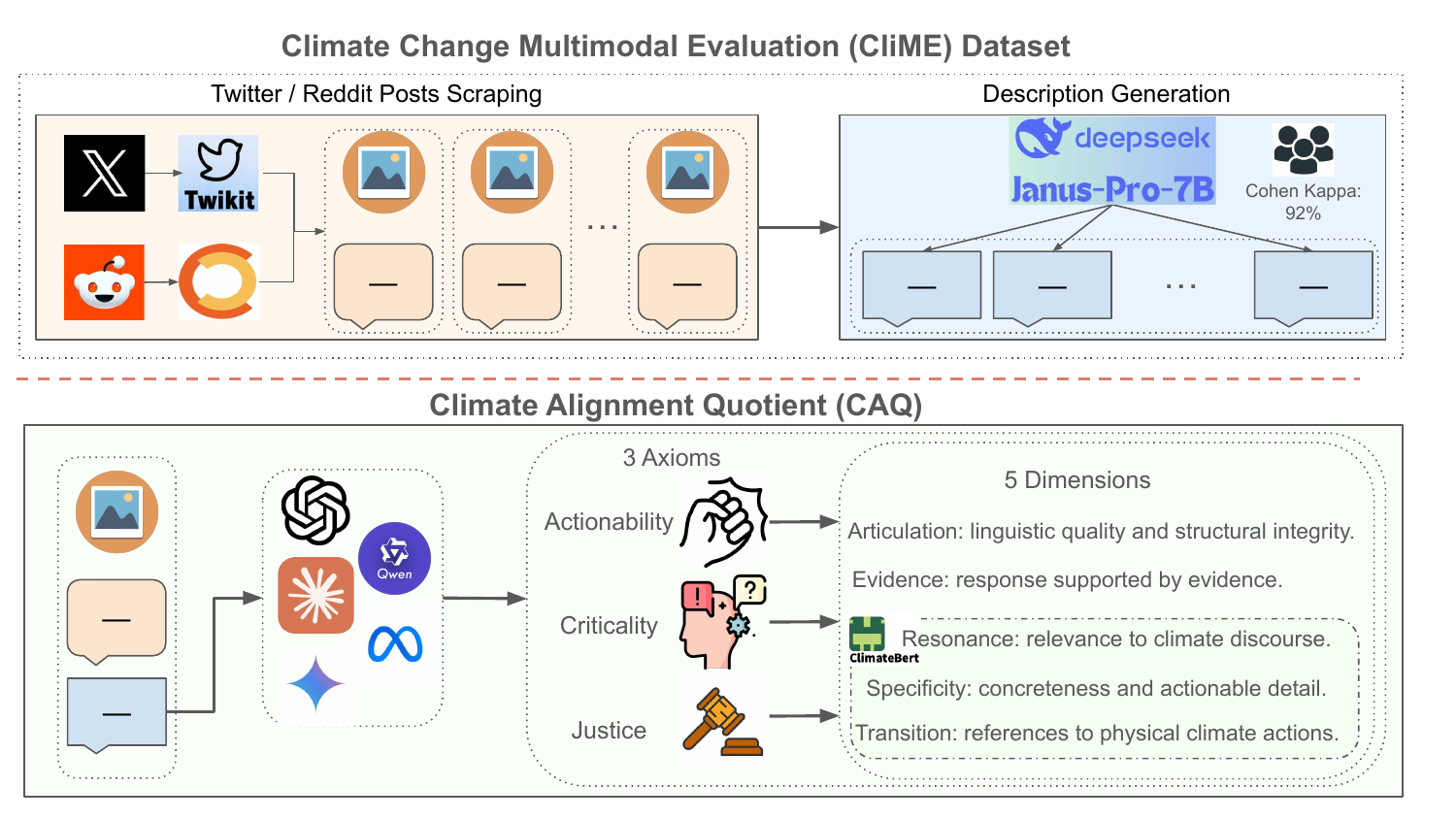}
    \vspace{-0.1cm}
    \caption{Overview of the \textit{Climate Change Multimodal Evaluation (CliME)} dataset and the \textit{Climate Alignment Quotient (CAQ)} workflow. The upper section illustrates the data collection process from Twitter and Reddit posts, utilizing multimodal sources (text and images) and description generation through the Janus-Pro-7B model followed by human annotations. The lower section demonstrates the CAQ evaluation framework, integrating multimodal data and analytical lenses (Actionability, Criticality, Justice) to assess climate communication across five dimensions: Articulation, Evidence, Resonance, Transition, and Specificity.}
    \label{fig:main_workflow}
    \vspace{-0.5cm}
\end{figure*}

To address these challenges and assess whether LLMs can interpret and generate credible climate communication, we introduce \textbf{CliME}, a benchmark comprising 2579 pairs of posts scraped from Reddit and Twitter, with most of the data focused on memes, infographics, and skeptic content. We then generate descriptions of these images and texts using DeepSeek Janus Pro \cite{chen2025janus} followed by human annotation (see Section \ref{genandhuman}). These descriptors serve as the basis for evaluating LLMs' capacity to comprehend and address climate change, guided by our proposed three key lenses: \textbf{(i) Actionability}, \textbf{(ii) Criticality}, and \textbf{(iii) Justice} (see Section \ref{analyticallens}). To assess the LLM responses generated through these lenses, we introduce the \textbf{Climate Alignment Quotient (CAQ)}, a novel metric that quantifies gaps across five critical axes: \textbf{(i) Articulation, (ii) Evidence,} \textbf{(iii) Resonance,} \textbf{(iv) Transition,} and \textbf{(v) Specificity} (see Section \ref{caq}), thereby determining the extent to which LLMs capture intrinsic climate knowledge. Figure \ref{fig:main_workflow} illustrates the entire workflow. In summary, our contributions are as follows:

(i) A first-of-its-kind multimodal benchmark, \textbf{CliME},  primarily featuring climate change related memes and skeptic content from social media.

(ii) Three analytical prompting paradigms: \textbf{Actionability, Criticality,} and \textbf{Justice} lenses, designed to investigate LLMs and assess their ability to interpret and generate credible climate discourse.

(iii) The \textbf{Climate Alignment Quotient (CAQ)}, a metric to measure the intrinsic alignment of LLMs in climate communication.

\section{Related Works}
\textbf{Climate Communication in Social Media.} \citealp{grundmann2010discourse} pioneered the analysis of climate change discourse through textual content, comparing climate-related word frequencies in news articles between Europe and the USA. With the rise of multimedia platforms like Twitter (X), Instagram, TikTok, WhatsApp, and YouTube, social media has been shown to enhance public awareness of climate issues \cite{farooq2024social}. \citealp{abdallah2023role} further demonstrated a positive correlation between climate-related social media content and increased public awareness. Studies highlight the effectiveness of personalized, relatable content in engaging audiences \citealp{leon2022social}, and social media is increasingly viewed as a trusted information source \citealp{hamed2023impact}. However, misinformation and echo chambers remain significant barriers to active engagement \citealp{abdallah2023role}. Hence, the analysis of climate change discourse on social media remains a rapidly evolving field that requires further attention. 

\textbf{LLMs in Climate Change Discourse Analysis.} Recent work has applied LLMs to analyze climate change discourse. ChatREPORT \cite{ni2023chatreport}, using ChatGPT \cite{hurst2024gpt} with expert-designed prompts, examined 9,781 corporate sustainability reports to evaluate climate action. \citealp{thulke2024climategpt} introduced ClimateGPT, a domain-specific LLM trained on 300 billion tokens (4.2 billion climate-related), validated through benchmarks and human evaluation. \citealp{zhou2024large} leveraged GPT-4 to uncover latent narratives in climate-related news from North American and Chinese sources.

While \cite{ni2023chatreport} and \cite{zhou2024large} adapts existing LLMs for climate discourse through prompt engineering, they assume the models' inherent capability to address climate issues. In contrast, \citealp{joe2024assessing} evaluated GPT-4o on climate tasks of varying expertise levels using GAMI literature \cite{berrang2021systematic}, revealing limitations in handling high-expertise tasks, especially those involving stakeholder identification and nuanced analysis \cite{hurst2024gpt}. There remains a gap in multi-perspective assessments of LLM responses across various climate change discourse sources, including news articles and social media.

\textbf{Multimodal Understanding in Climate Context.} Multimodal data is increasingly used in climate research for tasks like stance detection, predictive modeling, and video analysis \cite{dancygier2023multimodal, wang2024multiclimate, mohan2023multimodal, bai2024inferring, multi2024}. Social media platforms, especially Twitter (X) and Reddit, are key sources of such content, including humorous posts \cite{kovacheva2022climate, bai2024inferring}. Recent work has adapted vision-language models like CLIP \cite{radford2021learning} and BLIP \cite{li2022blip} to climate-specific tasks, achieving superior performance in stance detection and misinformation detection \cite{multi2024}. Benchmarks like MultiClimate \cite{wang2024multiclimate} and GreenScreen \cite{sharma2024greenscreen} evaluate models on image–text alignment, narrative coherence, and visual rhetoric in climate-skeptic content. However, they focus mainly on YouTube data, overlooking fast-spreading formats like memes and infographics.

\textbf{Assessment Frameworks for Climate Communication.} Three main frameworks have emerged for evaluating climate communication: (i) human-rated with AI support \cite{bulian2023assessing}, (ii) AI-based evaluation using ChatGPT \cite{gursesli2023chronicles}, and (iii) expert annotation with Likert scales \cite{nguyen2024my}. \citealp{bulian2023assessing} offers a qualitative approach, assessing text style, clarity, tone, and epistemological elements like accuracy and uncertainty. \citealp{gursesli2023chronicles} provides a quantitative method focused on narrative quality, measuring coherence, inspiration, and fluency. \citealp{nguyen2024my} presents a domain-specific framework using expert ratings to evaluate LLM-generated climate advice for agriculture. However, existing frameworks overlook climate communication via social media and humorous content.


\section{CliME Dataset}
\label{clime}
To understand how Large Language Models (LLMs) engage with and comprehend climate discourse, specifically in addressing climate change, we introduce the first-of-its-kind multimodal benchmark, \textbf{CliME\footnote{\url{https://huggingface.co/datasets/climedataset/CliME}} (Climate Change Multimodal Evaluation)} dataset. Comprising \textbf{2,579} data points, sourced and filtered from real-world Twitter and Reddit posts, primarily memes, skeptics, and infographics, CliME shifts the focus toward evaluating models based on their ability to generate credible, actionable, and equitable climate-related communication with multimodal (image-text) contexts.

\subsection{Dataset Creation}
\label{climecreation}
The CliME dataset was developed by systematically scraping posts from both Twitter (now X) and Reddit to capture a diverse spectrum of climate related discourse, including both memes and skeptical content. Data collection from X was conducted using Selenium and Twikit \cite{twikit}, enabling the extraction of posts tagged with climate-related hashtags such as \textbf{\#climatechange, \#climatememes, \#globalwarming, \#netzero, and \#climateskeptics}. Reddit data was obtained via the Yet Another Reddit Scraper (YARS) \cite{yars} library, specifically targeting posts and comments from climate-focused subreddits such as \textbf{r/climatememes} and \textbf{r/climateskeptics}. The initial raw dataset comprised approximately $\sim 4K$ posts, which subsequently underwent a rigorous filtering pipeline to enhance relevance and quality. First, language filtering was applied using the langdetect \cite{langdetect} library to exclude non-English texts, ensuring linguistic consistency for downstream analysis. Next, we conduct relevance verification with manual review to eliminate off-topic content, including unrelated memes and spam. Lastly, de-duplication was performed to remove redundant entries; perceptual hashing via ImageHash \cite{imagehash} was utilized to generate unique fingerprints for images, identifying and flagging near-identical visuals, such as reposted memes with minor modifications. After completing the filtering process, the final dataset comprised \textbf{2579} entries, each meeting the criteria for linguistic consistency, relevance, uniqueness, and quality.

\subsection{Descriptor Generation and Human Annotations}
\label{genandhuman}
A critical challenge lies in how LLMs process visual content: directly feeding raw pixels to text-based LLMs risks stripping away the nuanced, context-dependent narratives that images convey. This exploits the primal impact of visuals to lend credibility to false claims. To bridge this gap, we employ \textbf{DeepSeek Janus Pro}, an autoregressive Vision-Language Model (VLM) that unifies multimodal understanding and generation, to process both the text and its accompanying image, generating structured, context-aware descriptions. We configure DeepSeek with a temperature of 0.3 to balance specificity and creativity; max token length of 512. These detailed text descriptors help in assessing LLMs to distinguish fact from manipulation and ensure transparency in how visual context shapes their outputs. Subsequently, for robust verification, we manually annotated the generated descriptors. In this process, we carefully examined whether the combined interpretation of the post’s image and text aligned semantically with the descriptor’s intended meaning, assigning a score of 1 for a match and 0 for no match. The data was evaluated by two human annotators, and we filtered the data based on their evaluations. The inter-annotator agreement is 92.0\% measured by Cohen's kappa \cite{cohen1960cofficient}, showing near perfect agreement.
\vspace{-1pt}
Our final dataset comprises triples of image, original text, and generated descriptor, helping in systematic evaluation of assessing LLMs to distinguish fact from manipulation and ensure transparency in how visual context shapes their outputs (cf. Appendix \ref{moreclime} for examples).

\section{Climate Alignment Quotient (CAQ)}
\subsection{Analytical Lenses}
\label{analyticallens}
We evaluate LLMs' ability to interpret adversarial multimodal narratives, rebut misinformation with scientific rigor, and center justice for marginalized communities in climate descriptors through three novel analytical lenses: \textbf{(i) Actionability, (ii) Criticality, and (iii) Justice} (cf. Appendix \ref{lenses}). Descriptors are first passed through carefully crafted analytical prompts, which serve as interpretive lenses designed to assess how effectively LLMs engage with crucial climate discourse elements. For instance, consider the descriptor of a Twitter post as shown in Figure \ref{fig:main}: \textit{"The image and caption highlight the environmental policy of the Trump administration, specifically focusing on the withdrawal from the Paris Climate Agreement. The post emphasizes the risks associated with this decision, such as increased greenhouse gas emissions and the potential for climate-related disasters. It also highlights the Green New Deal as a potential solution, emphasizing the need for a comprehensive and immediate response to climate change. The post encourages action and emphasizes the urgency of addressing climate issues."} Our analytical lenses would evaluate this as follows:

\textbf{Actionability} assesses the translation of climate discourse into concrete interventions. In our example, the post's endorsement of the \textit{"Green New Deal"} as a solution is scrutinized for actionable components. While the proposal advocates for broad objectives like transitioning to \textit{"renewable energy"}, its practical feasibility may be rated as medium due to potential political gridlock and ambiguous funding mechanisms. The message emphasizes urgency but lacks specifics, such as identifying legislative bodies that would \textit{champion the policy}, setting deadlines for key milestones like grid \textit{"de-carbonization"}, or outlining strategies to mobilize workforce training programs. Unaddressed gaps include potential economic strains on industries reliant on fossil fuels and the absence of contingency plans for technological bottlenecks, revealing a disconnect between aspirational goals and practical roadmaps.

\textbf{Criticality} examines the structural roots and foundations of climate narratives. The post links the Trump administration's withdrawal from the \textit{"Paris Agreement"} to increased emissions and climate related disasters. A critical analysis would question this causal oversimplification, noting that while the withdrawal weakened global cooperation, emissions were already rising due to entrenched systems like \textit{"fossil fuel"} lobbying and inadequate clean energy incentives. Framing the Green New Deal as a singular solution might sidestep debates over its scalability, such as conflicting estimates about job creation or its silence on nuclear energy's role. This uncritical portrayal risks reinforcing partisan divides rather than addressing systemic barriers like corporate influence on \textit{climate policy}.

\textbf{Justice} centers on marginalized voices and systemic inequities. The post's U.S. (United States)-centric focus overlooks how the withdrawal from the Paris Agreement undermined climate financing for \textit{"Global South"} nations, exacerbating vulnerabilities in regions least responsible for emissions. By positioning the Green New Deal as a domestic fix, the narrative neglects historical U.S. accountability for global emissions and fails to address how transition costs might disproportionately affect low income communities, such as rising energy prices or displacement from renewable infrastructure projects. A justice lens would highlight absent voices, such as Indigenous groups advocating for land sovereignty in solar farm expansions, and question whether the policy redistributes power or perpetuates existing inequities.

Following evaluation through these analytical lenses, the resulting outputs provide essential context and data to progress into our quantitative evaluation step: calculating the \textbf{Climate Alignment Quotient (CAQ)}.

\subsection{CAQ}
\label{caq}
The \textbf{Climate Alignment Quotient (CAQ)} is a composite metric specifically designed to quantify the effectiveness and alignment of climate-related communication generated by LLMs. Integrating and systematically evaluating outputs derived from the analytical lenses (Actionability, Criticality, Justice), the CAQ assesses alignment across five critical dimensions: \textbf{(i) Articulation, (ii) Evidence, (iii) Resonance, (iv) Transition, and (v) Specificity}. The CAQ score is calculated as a weighted sum of the five core metrics, mathematically can be described as:
\begin{align}
\textbf{CAQ} &= w_1 \cdot \textit{Resonance} + w_2 \cdot \textit{Articulation} \notag \\
&\quad + w_3 \cdot \textit{Evidence} + w_4 \cdot \textit{Transition} \notag \\
&\quad + w_5 \cdot \textit{Specificity}
\end{align}

where $w_1 + w_2 + w_3 + w_4 + w_5$ aggregates to 1. Empirically, we set $w_1 = 0.25$, $w_2 = 0.3$, $w_3 = 0.2$, $w_4 = 0.15$, and $w_5 = 0.1$.

\textbf{(i) Articulation:} Our articulation score measures the linguistic quality and structural integrity of climate communications through a dual component analysis. The articulation score is calculated as an equally weighted combination of coherence and completeness, given by:

\begin{equation}
\textit{Articulation} = 0.5 \cdot \textit{Coherence} + 0.5 \cdot \textit{Completeness}
\end{equation}

The \textit{coherence} component assesses how well sentences connect and flow together and is further broken down into:

\begin{equation}
\textit{Coherence} = 0.6 \cdot \textit{Syntactic} + 0.4 \cdot \textit{Semantic}
\end{equation}

\textit{Syntactic} coherence evaluates the presence of discourse markers (coordinating conjunctions, subordinating conjunctions, and tokens with "mark" dependency) relative to sentence count. This provides a quantitative measure of how well the text is structurally connected through explicit linguistic devices. \textit{Semantic} coherence, computed using the \texttt{all-mpnet-base-v2} Sentence Transformer \cite{song2020mpnet} model, measures thematic consistency between adjacent sentences through embedding similarity, capturing the continuity of ideas throughout the text. The completeness component evaluates the grammatical integrity of text by assessing syntactic structure at the sentence level. We identify two types of valid constructions: complete statements containing both subject and predicate elements, and imperative sentences beginning with action verbs. Using SpaCy's \cite{spacy2} dependency parsing, we identify subjects through dependency relations (\texttt{"nsubj"} or \texttt{"nsubjpass"}) and predicates via ROOT verbs. The \textit{completeness} score represents the proportion of sentences that follow either of these grammatical patterns relative to the total sentence count. This measure effectively captures whether climate communications maintain consistent grammatical structure throughout, ensuring clarity of the conveyed information by the model.

\textbf{(ii) Evidence:} The Evidence metric distinctly assesses the extent to which the climate communications in the LLM’s responses are substantiated by verifiable data and concrete examples. We employ the specialized NLP model \textit{"climate-nlp/longformer-large-4096-1-detect-evidence"} \cite{NEURIPS2023_7ccaa4f9} to compute this score. A high Evidence score indicates that the language model’s claims are supported by detailed, traceable evidence, such as explicit data points, clear references, and documented instances of climate action demonstrating a robust and genuine commitment to addressing climate challenges. In contrast, a low score reveals that the communications lack sufficient backing, potentially suggesting superficial engagement or greenwashing.

The remaining three CAQ components are derived via a suite of models from ClimateBERT \cite{webersinke2021climatebert}, pretrained on a text corpus comprising climate related research paper abstracts, corporate and general news and reports from companies. The measures derived from this collection are as follows:

\textbf{(iii)} \textbf{Resonance:} The Resonance metric leverages the 
\textit{"climatebert/distil\-roberta-base-climate-detector"} model to quantify how strongly content engages with climate context. This model performs binary classification, outputting a probability score that represents the likelihood of climate relevance. Higher scores indicate content that more directly addresses climate change concepts, terminology, and themes. The detector performs well at identifying subtle climate references while filtering out environmental content unrelated to climate change.

\textbf{(iv)} \textbf{Transition:} The Transition metric uses the \textit{"climatebert/transition-physical"} model to evaluate references to physical climate transition processes. It identifies content related to tangible climate adaptation and mitigation practices, infrastructure changes, and transitions in physical systems. It detects mentions of renewable energy implementation, carbon capture technologies, climate-resilient infrastructure, and other physical interventions. Higher scores indicate content that addresses concrete transitional mechanisms rather than abstract climate concepts or general environmental concerns.

\textbf{(v)} \textbf{Specificity:} The Specificity metric utilizes the \textit{"climatebert/distilroberta-base-climate-specificity"} model to assess whether given response by the model provides specific, actionable information versus general statements or vague recommendations. Higher specificity scores indicate content containing concrete actions, measurable targets, defined timelines, or detailed examples. This dimension is crucial for distinguishing between aspirational climate rhetoric and communications that provide implementable guidance or precise information that can drive meaningful action in real life.

In summary, CAQ serves as a robust metric for evaluating the effectiveness of climate communications in LLMs. Higher CAQ scores indicate communications that strongly align with climate objectives, characterized by coherent articulation, high resonance, strong evidence, concrete transitional strategies, and actionable specificity; hence, aligning communication strategies of LLMs with climate objectives.

\begin{table*}[t]
\centering
\caption{Climate Alignment Quotient (CAQ) Comparison Across Different LLMs. 
Color intensity indicates performance level, with deeper colors representing higher scores. 
Column headers are color-coded by lens (yellow for Actionability, red for Criticality, 
green for Justice), while component scores use blue shades proportional to their values.}
\label{tab:caq-comparison}
\setlength{\tabcolsep}{2.5pt}
\renewcommand{\arraystretch}{1.2}
\begin{tabular}{l|ccccc|c|ccccc|c|ccccc|c}
\toprule
\rowcolor{gray!15}
& \multicolumn{6}{c|}{\cellcolor{yellow!15}\textbf{Actionability}} & \multicolumn{6}{c|}{\cellcolor{red!15}\textbf{Criticality}} & \multicolumn{6}{c}{\cellcolor{green!15}\textbf{Justice}} \\
\cmidrule{2-19}
\rowcolor{gray!15}
\textbf{Model} & \rot{\textbf{R}} & \rot{\textbf{A}} & \rot{\textbf{E}} & \rot{\textbf{T}} & \rot{\textbf{Sp}} & \textbf{CAQ} & \rot{\textbf{R}} & \rot{\textbf{A}} & \rot{\textbf{E}} & \rot{\textbf{T}} & \rot{\textbf{Sp}} & \textbf{CAQ} & \rot{\textbf{R}} & \rot{\textbf{A}} & \rot{\textbf{E}} & \rot{\textbf{T}} & \rot{\textbf{Sp}} & \textbf{CAQ} \rule[-5pt]{0pt}{20pt} \\
\midrule
GPT-4o & \cellcolor{blue!20}1.0 & \cellcolor{blue!15}.86 & \cellcolor{blue!10}.68 & \cellcolor{blue!8}.19 & \cellcolor{blue!5}.18 & \cellcolor{purple!20}\textbf{.69} & 
\cellcolor{blue!20}1.0 & \cellcolor{blue!10}.81 & \cellcolor{blue!15}.85 & \cellcolor{blue!5}.15 & \cellcolor{blue!8}.18 & 
\cellcolor{purple!35}\textbf{.70} & 
\cellcolor{blue!20}1.0 & \cellcolor{blue!10}.81 & \cellcolor{blue!15}.98 & \cellcolor{blue!8}.18 & \cellcolor{blue!5}.15 & \cellcolor{purple!35}\textbf{.73} \\

\rowcolor{gray!5}
Claude 3.7 Sonnet & \cellcolor{blue!20}1.0 & \cellcolor{blue!10}.82 & \cellcolor{blue!15}.95 & \cellcolor{blue!5}.18& \cellcolor{blue!8}.20 & \cellcolor{purple!50}\textbf{.73} & 
\cellcolor{blue!20}1.0 & \cellcolor{blue!10}.84 & \cellcolor{blue!15}.97 & \cellcolor{blue!5}.17 & \cellcolor{blue!8}.21 & \cellcolor{purple!50}\textbf{.74} & 
\cellcolor{blue!20}1.0 & \cellcolor{blue!15}.83 & \cellcolor{blue!20}1.0 & \cellcolor{blue!8}.11& \cellcolor{blue!10}.18 & \cellcolor{purple!35}\textbf{.73} \\

Gemini 2.0 Flash & \cellcolor{blue!20}1.0 & \cellcolor{blue!10}.74 & \cellcolor{blue!15}.84 & \cellcolor{blue!8}.18 & \cellcolor{blue!5}.15 & \cellcolor{purple!15}\textbf{.68} & 
\cellcolor{blue!20}1.0 & \cellcolor{blue!10}.69 & \cellcolor{blue!15}.78 & \cellcolor{blue!5}.17 & \cellcolor{blue!8}.20 & \cellcolor{purple!20}\textbf{.66} & 
\cellcolor{blue!20}1.0 & \cellcolor{blue!10}.82 & \cellcolor{blue!15}.99 & \cellcolor{blue!5}.14 & \cellcolor{blue!8}.15 & \cellcolor{purple!35}\textbf{.73} \\

\rowcolor{gray!5}
LLaMA 3.3 70b & \cellcolor{blue!20}1.0& \cellcolor{blue!10}.72 & \cellcolor{blue!15}.74 & \cellcolor{blue!8}.21 & \cellcolor{blue!5}.15 & \cellcolor{purple!10}\textbf{.66} & 
\cellcolor{blue!20}1.0 & \cellcolor{blue!10}.78 & \cellcolor{blue!15}.86 & \cellcolor{blue!5}.14 & \cellcolor{blue!8}.19 & \cellcolor{purple!35}\textbf{.70} & 
\cellcolor{blue!20}1.0 & \cellcolor{blue!15}.84 & \cellcolor{blue!20}1.0 & \cellcolor{blue!8}.15 & \cellcolor{blue!10}.19 & \cellcolor{purple!50}\textbf{.74} \\

Qwen QwQ 32b & \cellcolor{blue!20}.99& \cellcolor{blue!15}.81 & \cellcolor{blue!10}.80& \cellcolor{blue!8}.20 & \cellcolor{blue!8}.20 & \cellcolor{purple!35}\textbf{.70} & 
\cellcolor{blue!20}0.99 & \cellcolor{blue!10}.77 & \cellcolor{blue!15}.86 & \cellcolor{blue!8}.17 & \cellcolor{blue!5}.22 & \cellcolor{purple!35}\textbf{.70} & 
\cellcolor{blue!20}1.0 & \cellcolor{blue!15}.77 & \cellcolor{blue!20}1.0 & \cellcolor{blue!10}.19 & \cellcolor{blue!8}.18 & \cellcolor{purple!35}\textbf{.73} \\
\bottomrule
\end{tabular}

\vspace{2mm}
\begin{center}
\footnotesize{Legend: R = Resonance, A = Articulation, E = Evidence, T = Transition, Sp = Specificity} \\
\footnotesize{Weights: Resonance (0.25), Articulation (0.30), Evidence (0.20), Transition (0.15), Specificity (0.1)}
\end{center}
\end{table*}

\subsection{Evaluation}
\begin{figure}[ht]
    \centering
    \includegraphics[width=0.5\textwidth]{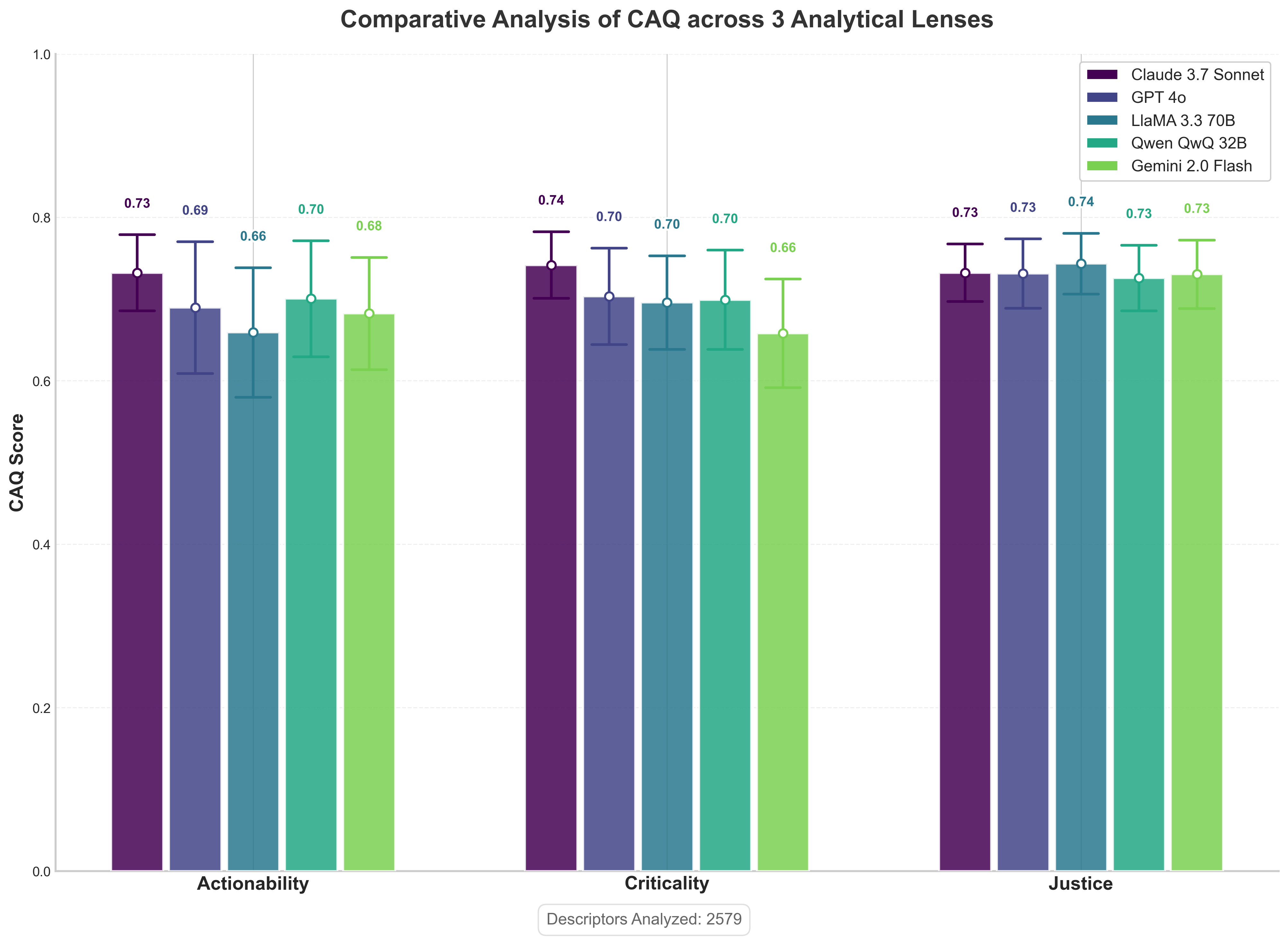}
    \caption{
    Comparative analysis of CAQ scores across the \textit{Actionability}, \textit{Criticality}, and \textit{Justice} lenses for five large language models: Claude 3.7 Sonnet, GPT-4o, LLaMA 3.3 70B, Qwen QwQ 32B, and Gemini 2.0 Flash, on CliME. Each bar represents the mean CAQ score, and error bars indicate the standard deviation, showcasing the variability in model performance. \texttt{Claude 3.7 Sonnet} is seen to generally outperform other models, across all the lenses, with scores consistently above 0.70 and relatively consistent standard deviations.
    }
    \label{fig:models_analysis}
\end{figure}

We benchmarked five state-of-the-art LLMs on \text{CliME} and evaluated them using our CAQ metric. The models include GPT-4o \cite{hurst2024gpt}, LLaMA 3.3 70B \cite{dubey2024llama}, Gemini 2.0 Flash \cite{team5gemini}, Qwen QwQ (Qwen with Question) 32B \cite{yang2024qwen2}, and Claude 3.7 Sonnet \cite{claudesonnet}. All models were configured to generate outputs at a temperature of 0.1 to ensure deterministic responses. The scores are reported in Table~\ref{tab:caq-comparison}; a comprehensive analysis of the CAQ scores across the models along our proposed lenses is shown in Figure~\ref{fig:models_analysis}. We find from our experiments that resonance (R) scores are consistently high (near 1.0) across all models, indicating a strong alignment with climate-related context in the generated descriptors. In terms of \textit{Actionability}, Claude 3.7 Sonnet achieves an average CAQ of 0.73, while Gemini 2.0 Flash and LLaMA 3.3 70B record slightly lower scores of 0.68 and 0.66, respectively; GPT-4o and Qwen both score around 0.70. The \textit{Criticality} lens shows Claude and LLaMA performing comparably (0.74 and 0.70, respectively), with Gemini 2.0 Flash trailing at 0.66. For the \textit{Justice} dimension, all models converge between 0.73 and 0.74. This uniformity in Justice scores suggests that fairness and equity considerations are consistently addressed, while the variability in Actionability and Criticality highlights differences in the models' abilities to generate concrete, actionable guidance and critically evaluative discourse. The articulation (A) measure further reveals that GPT-4o (0.86) and Claude 3.7 Sonnet (0.82) offer more coherent communication compared to Gemini 2.0 Flash (0.74) and LLaMA 3.3 70B (0.72). Moreover, the evidence (E) scores are particularly high for Claude 3.7 Sonnet and LLaMA 3.3 70B, indicating that these models more effectively ground their outputs with verifiable data. In addition to this, transition scores are consistently low (ranging from 0.11 to 0.21) across all models, suggesting a common challenge in referencing physical climate transition processes. Overall, while all models demonstrate a balanced understanding of climate discourse with overall CAQ scores hovering between 0.70 and 0.74, Claude 3.7 Sonnet and GPT-4o tend to produce more articulate and evidence-supported discourse, whereas Gemini 2.0 Flash shows slightly lower performance overall.

\subsection{Analysis}
In this section, we present a comprehensive analysis of the \textbf{CAQ} scores across the proposed lenses of \textbf{Actionability}, \textbf{Criticality}, and \textbf{Justice}. This analysis encompasses both the distribution of individual CAQ scores and the evaluation of gaps between these dimensions, providing a detailed understanding of the multimodal climate discourse generated by LLMs.
\subsubsection{Distribution of CAQ Scores}
\begin{figure}[ht]
    \centering
    \includegraphics[width=0.45\textwidth]{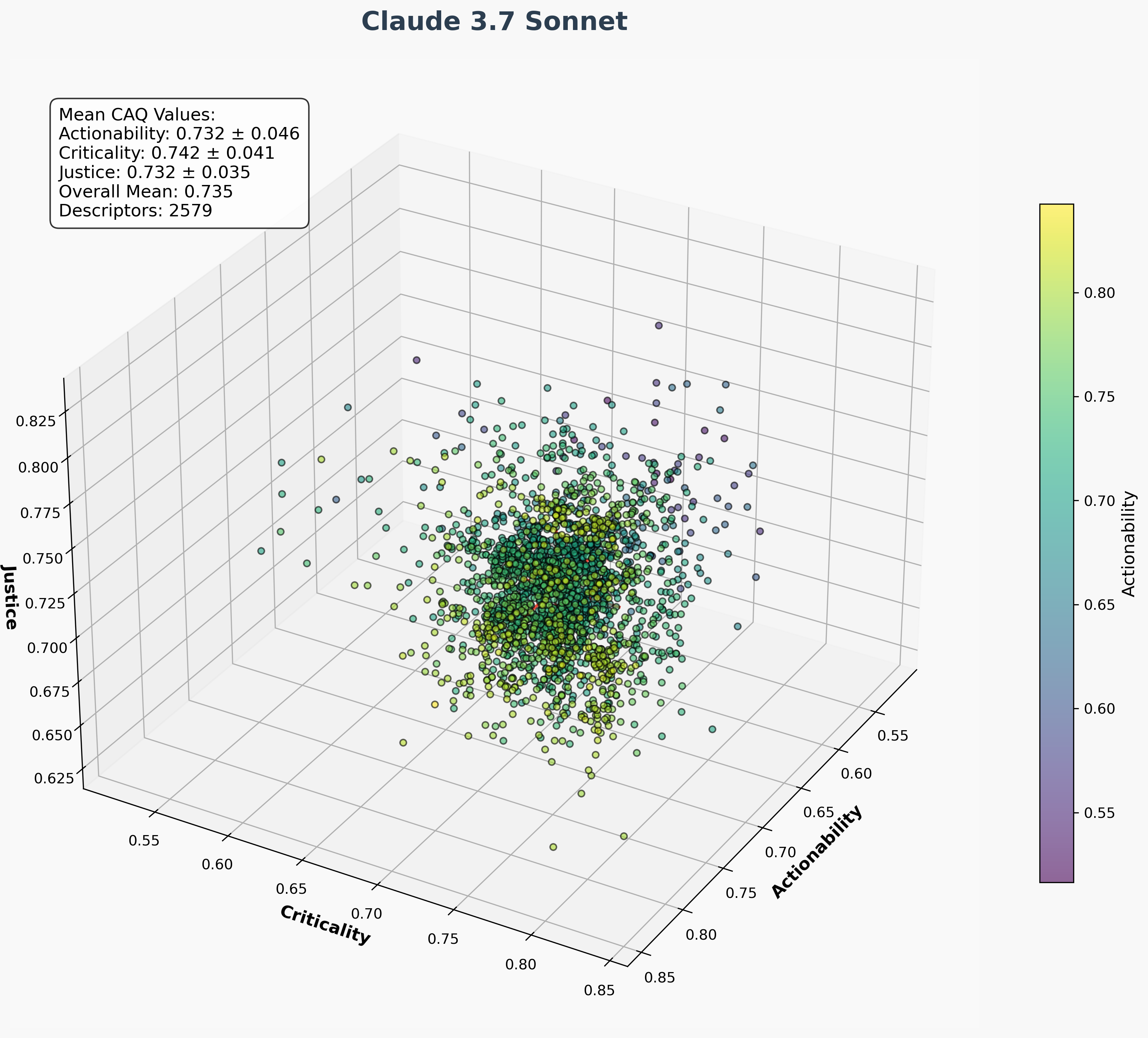}
    \caption{
    3D scatter plot of CAQ scores for the \texttt{Claude 3.7 Sonnet} model on the \text{CliME} dataset. Each data point represents a description's CAQ values along three axes: \textit{Actionability} (x-axis), \textit{Criticality} (y-axis), and \textit{Justice} (z-axis). The color scale (legend on the right) indicates the CAQ score of Actionability. Points near the center denote balanced discourse across all dimensions, whereas deviations along any axis suggest an overemphasis or under-representation of that particular lens.
    }
    \label{fig:3d_distribution}
\end{figure}

In Figure~\ref{fig:3d_distribution}, where the x-axis is \textit{Criticality}, the y-axis is \textit{Justice}, and the z-axis is \textit{Actionability}, for \texttt{Claude 3.7 Sonnet} (cf. Appendix \ref{3d} for other models), we observe that most points cluster around the mid-range of each axis (approximately 0.70-0.80), indicating that the LLM-generated content tends to balance urgency (Criticality), fairness (Justice), and practicality (Actionability). When \textit{Actionability} increases (both in the z-coordinate and in the color scale), there is often a slight upward score in both \textit{Criticality} and \textit{Justice}, suggesting that more action-oriented content tends to incorporate at least moderate levels of urgency and equity considerations. Conversely, at lower \textit{Actionability} scores (cooler colors), there's a wider spread in \textit{Criticality} and \textit{Justice}, suggesting that content with fewer calls to action is perceived with more varied urgency and equity.

\subsubsection{Gap Analysis}
In addition to this, we perform a gap analysis to evaluate the differences between CAQ scores across our proposed axes (cf. Appendix \ref{moregap} for box plots and heatmaps for all the models). Effective climate communication necessitates a balanced integration of these aspects. Overemphasis on one dimension, such as \textit{Actionability}, at the expense of others like \textit{Justice}, can lead to skewed narratives that overlook systemic inequities or fail to motivate comprehensive action.
In the case of \texttt{Claude 3.7 Sonnet}, one of the best scoring models on CAQ, one of the best-scoring models on CAQ, we empirically find that the average gap between dimensions varies across the 2,579 evaluated descriptors. The gap between \textit{Criticality} and \textit{Justice} is the smallest at 0.0313, 
showing these dimensions are generally well-aligned. However, the gap between \textit{Actionability} and \textit{Justice} tends to be larger, averaging around 0.0344, suggesting a slight under representation of justice considerations in certain outputs. Similarly, the \textit{Actionability-Criticality} gap averages about 0.0324. In case of Claude 3.7 Sonnet, our dimensional frequency analysis 
reveals that the most common largest gap type is between 
\textit{Actionability} and \textit{Criticality}, occurring in 37.1\% of descriptors. These variations across different posts provide valuable insights for developing more holistic approaches to climate communication that effectively balance considerations across these three dimensions.


\section{Conclusion}
In this work, we present \textit{CliME}, a novel multimodal dataset curated from social media, primarily featuring memes and skeptics, and introduce the \textit{Climate Alignment Quotient (CAQ)} as a comprehensive metric to evaluate climate discourse generated by LLMs. By benchmarking five state-of-the-art models, we analyzed their outputs across three lenses: \textit{Actionability, Criticality, and Justice}. Our experiments reveal that while all models consistently capture climate-related context, significant variations exist in generating actionable and critically evaluative messages. The gap analysis uncovers subtle imbalances among these dimensions, indicating areas for improvement in the models’ outputs. In today's landscape, misinformation and polarized narratives on climate change contribute to social harm and undermine effective action; our CAQ framework offers a promising approach to understanding the strengths and weaknesses of LLMs. By fostering balanced climate communication, our work aims to prevent the spread of misleading information and support informed public dialogue, ultimately guiding policy-making for social good. Future efforts will focus on refining the CAQ metric and expanding the \textit{CliME} dataset to include a broader spectrum of multimodal content, thereby empowering both LLMs and VLMs to contribute more effectively to a sustainable future.

\section*{Limitations}
Though the proposed CAQ framework provides a structured and comprehensive way to evaluate climate discourse, it currently relies on existing pre-trained models for its assessment. Since these models may not be extensively trained on the latest social media data, particularly memes and other highly contextual content, there is a risk of missing nuanced climate change signals. Additionally, the nature of social media platforms, where language evolves rapidly and memes can quickly become outdated or repurposed, presents challenges in ensuring that all relevant domain-specific shifts are captured.

\section*{Ethics Statement}
All data in our \text{CliME} dataset originate from publicly accessible Reddit and Twitter posts. We strictly followed platform guidelines during data collection, focusing on content explicitly marked for public sharing and ensuring that no personally identifiable information (PII) was retained. Although memes, infographics, and other materials often exhibit strong emotional or political underpinnings, our goal is to assess climate-related discourse rather than endorse any particular viewpoint.
The proprietary models employed in our study were accessed strictly via valid subscriptions, in accordance with the terms of service provided by the respective providers.

\bibliography{main}

\appendix
\section{Appendix}
\subsection{Analytical Lenses}
\label{lenses}
We present the prompts used for our three proposed lenses: \textit{Actionability}, \textit{Criticality}, and \textit{Justice}.

\begin{tcolorbox}[colback=yellow!10, colframe=purple!90, title=Actionability]
\textbf{Description:} \{description\} \\[5pt]
\textbf{Instruction:} Analyze the climate-related message in the above description through an actionability lens. Respond in one unified paragraph that summarizes the key climate issues, identifies actionable solutions, evaluates their feasibility (high/medium/low), assesses explicit commitments (who, what, when, how), and highlights risks or unaddressed challenges. Do not output any extra information other than this analysis in your response.
\end{tcolorbox}

\begin{tcolorbox}[colback=blue!10, colframe=blue!50, title=Criticality]
\textbf{Description:} \{description\} \\[5pt]
\textbf{Instruction:} Critically evaluate the climate-related message in the above description. Respond in one unified paragraph that identifies core claims, assesses evidence quality, highlights unsubstantiated claims or oversimplified arguments, evaluates engagement with competing viewpoints, and analyzes its impact on climate discourse. Do not output any extra information other than this analysis in your response.
\end{tcolorbox}

\begin{tcolorbox}[colback=green!10, colframe=green!50, title=Justice]
\textbf{Description:} \{description\} \\[5pt]
\textbf{Instruction:} Analyze the climate-related message in the above description through a justice lens. Respond in one unified paragraph that identifies centered/absent communities, assesses distribution of responsibility, evaluates acknowledgment of historical power imbalances, examines impacts on marginalized groups, and considers inclusion of cultural contexts. Do not output any extra information other than this analysis in your response.
\end{tcolorbox}
\subsection{CliME Examples}
\label{moreclime}
Table~\ref{table:examples_table} presents various Twitter and Reddit posts, alongside their corresponding images, textual content, and the automatically generated descriptions produced by Janus Pro filtered after human annotations.
\begin{table*}[h]
\centering
\scalebox{0.7}{
\begin{tabular}{m{1cm}|m{10cm}|m{10cm}} 
    \toprule
    \centering No. & \centering \textbf{Reddit/Twitter Post} & \centering \textbf{Description} \tabularnewline
    \midrule
    \centering\arraybackslash 1 &  
    \begin{minipage}[t]{10cm}
        \vspace{-2mm} 
        \centering
        \includegraphics[scale=0.95]{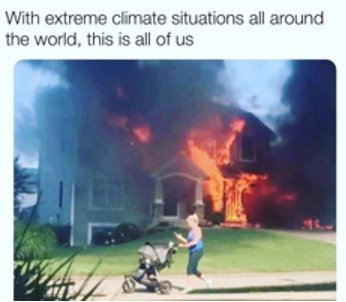} \\[-2mm] 
        So true it hurts :'$\textunderscore$
    \end{minipage} & 
    \begin{minipage}[t]{10cm}
        \vspace{-2mm} 
        The image and caption highlight the severe impact of climate change on communities, particularly vulnerable populations. The burning house in the background symbolizes the increasing frequency and intensity of natural disasters, such as wildfires, which are exacerbated by climate change. The caption emphasizes the shared responsibility of addressing these challenges, urging collective action to mitigate the risks and protect communities. ...
    \end{minipage} \tabularnewline
    \midrule
    \centering\arraybackslash 2 &  
    \begin{minipage}[t]{10cm}
        \vspace{-2mm} 
        \centering
        \includegraphics[scale=0.95]{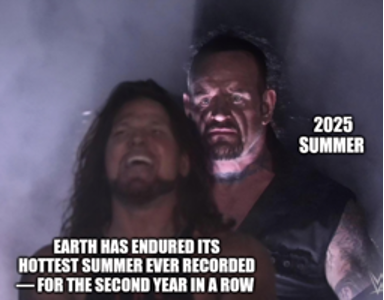} \\[-2mm] 
        And 2026 and 2027 and so on ...
    \end{minipage} & 
    \begin{minipage}[t]{10cm}
        \vspace{-2mm} 
        The image and caption highlight the alarming reality of Earth's climate, specifically the record-breaking heat experienced during the summer of 2025. The caption emphasizes the urgency of addressing climate change, stating that the second year in a row has been the hottest on record. It underscores the risks associated with climate change, such as extreme heat events, and calls for actionable solutions to mitigate these risks. The image of two individuals, one looking distressed and the other looming ominously, symbolizes the dire consequences of inaction and the need for immediate and collective action to protect the planet.
    \end{minipage} \tabularnewline
    \midrule
    \centering\arraybackslash 3 &  
    \begin{minipage}[t]{10cm}
        \vspace{-2mm} 
        \centering
        \includegraphics[scale=0.95]{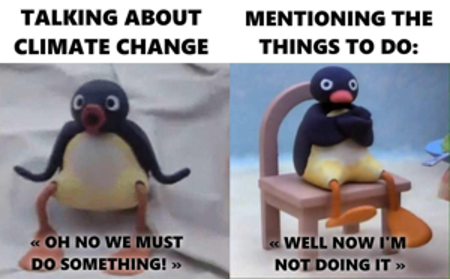} \\[-2mm] 
        Meat/car/air travel are so taboo
    \end{minipage} & 
    \begin{minipage}[t]{10cm}
        \vspace{-2mm} 
        The image and caption highlight the urgent need for action in addressing climate change. The left side of the image shows a character expressing concern and urgency, with the text "OH NO WE MUST DO SOMETHING!!" emphasizing the need for immediate action. The right side of the image shows a character sitting with crossed arms, indicating a lack of action, with the text "WELL NOW I'M NOT DOING IT." This contrast underscores the importance of taking concrete steps to mitigate climate change. ...
    \end{minipage} \tabularnewline
    \midrule
    \centering\arraybackslash 4 &  
    \begin{minipage}[t]{10cm}
        \vspace{-2mm} 
        \centering
        \includegraphics[scale=0.95]{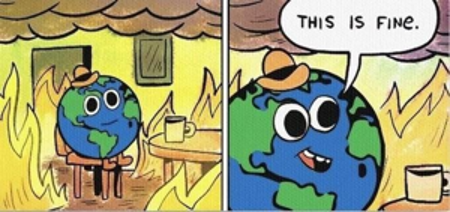} \\[-2mm] 
        What a gas! How serious climate action can be funny, too
    \end{minipage} & 
    \begin{minipage}[t]{10cm}
        \vspace{-2mm} 
        The image depicts a cartoon of the Earth, anthropomorphized and wearing a hat, sitting in a room engulfed in flames. The Earth character appears calm and unfazed, sipping from a mug. The caption reads, "This is fine."
        The climate-related message in this post is a satirical commentary on the lack of urgency and action in addressing climate change. The phrase "This is fine" is often used to dismiss serious issues, and in this context, it highlights the complacency and denial that can hinder meaningful climate action...
    \end{minipage} \tabularnewline
    \midrule
    \centering\arraybackslash 5 &  
    \begin{minipage}[t]{10cm}
        \vspace{-2mm} 
        \centering
        \includegraphics[scale=0.95]{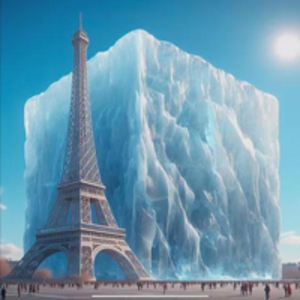} \\[-2mm] 
        What 30 million tonnes of ice that Greenland loses every hour looks like
    \end{minipage} & 
    \begin{minipage}[t]{10cm}
        \vspace{-2mm} 
        The image depicts a massive iceberg, approximately 30 million tonnes in size, floating in the Arctic Ocean. This visual representation highlights the alarming rate at which Greenland is losing ice, with an estimated 100 billion tonnes of ice melting every hour. The caption emphasizes the urgent need for climate action, as the loss of ice threatens global sea levels, ecosystems, and human societies ... \newline In conclusion, the post serves as a powerful reminder of the urgent need for climate action and emphasizes the importance of taking immediate and collective steps to address the climate crisis.
    \end{minipage} \tabularnewline
    \bottomrule

\end{tabular}}
\caption{More data samples from CliME.}
\label{table:examples_table}
\end{table*}

\subsection{Gap Analysis}
\begin{figure*}[ht]
\centering
\includegraphics[width=\textwidth]{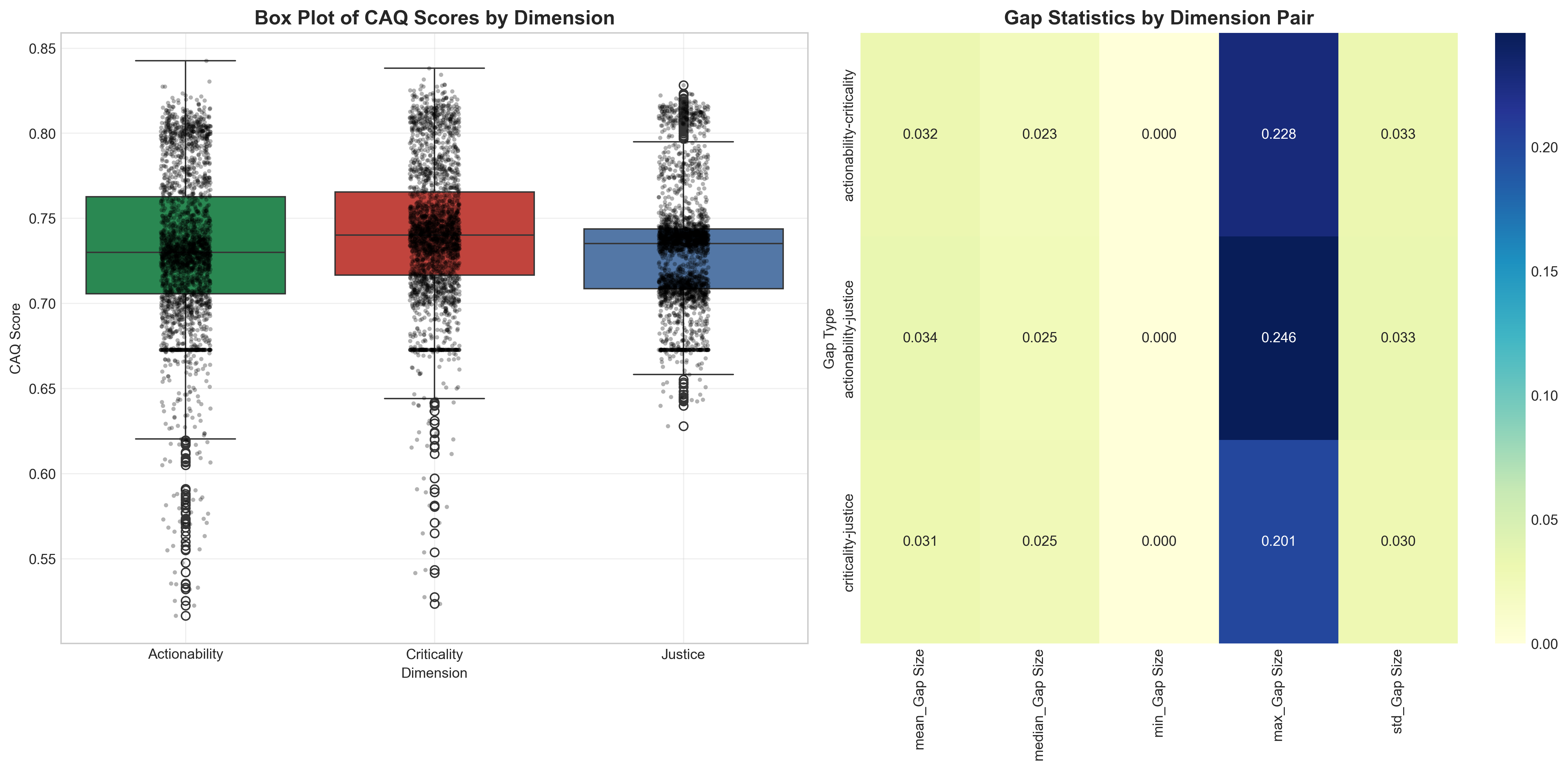}
\caption{Gap Analysis for \texttt{Claude 3.7 Sonnet}'s CAQ Score Performance. The left panel shows box plots of scores across three dimensions: Actionability (mean: 0.7321), Criticality (mean: 0.7416), and Justice (mean: 0.7321). The right panel displays a heatmap of gap statistics between dimension pairs, with the Actionability-Justice gap (0.0344) being the most significant, followed by the Actionability-Criticality gap (0.0324), while the Criticality-Justice gap (0.0313) shows the best balance. The analysis reveals more balanced dimensional scores compared to other models, with fewer large gaps across all dimension pairs.}
\label{fig1}
\end{figure*}

\begin{figure*}[ht]
\centering
\includegraphics[width=\textwidth]{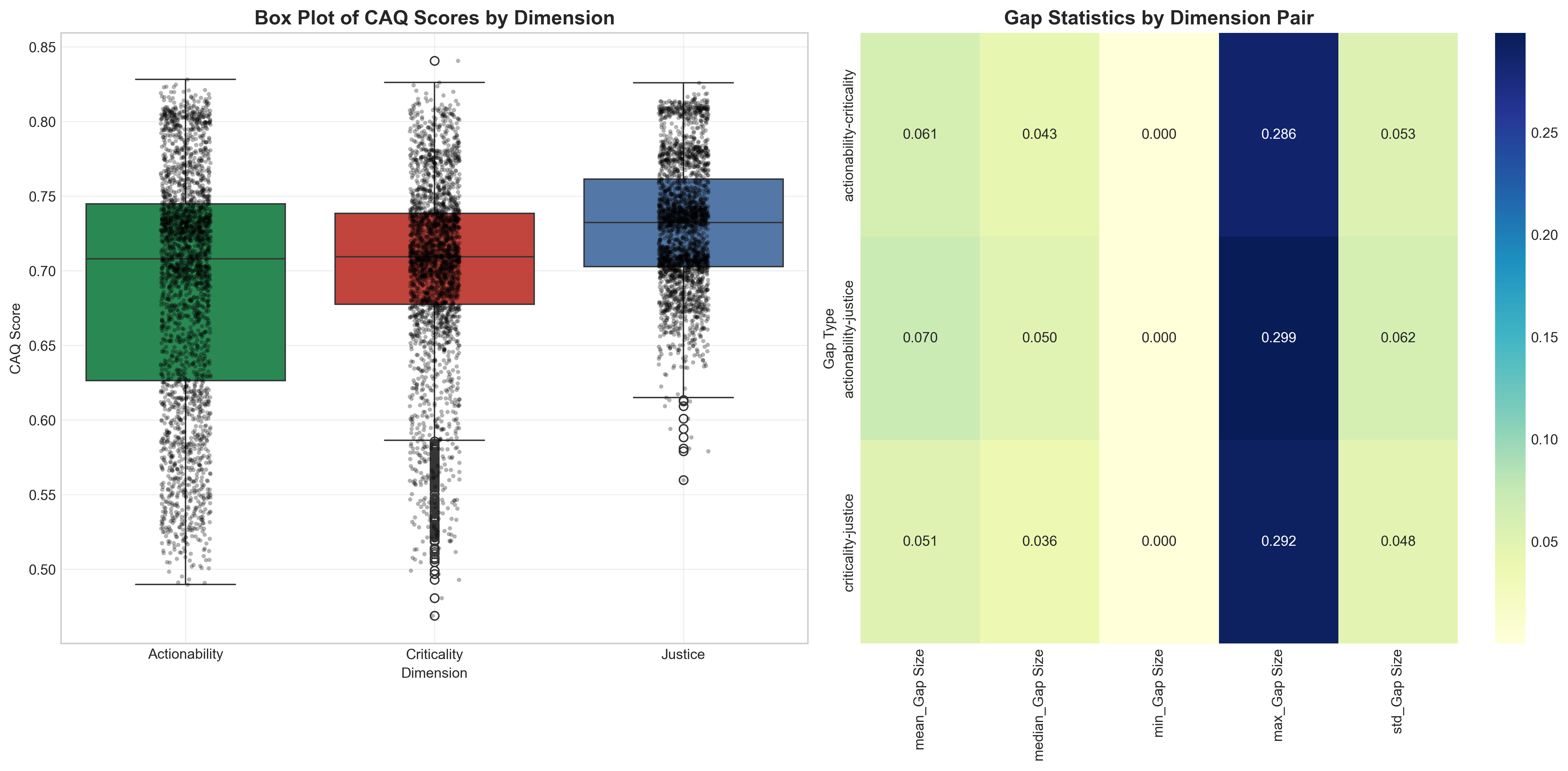}
\caption{
Gap Analysis for \texttt{GPT-4o}'s CAQ scores based on 2579 descriptors. The left panel shows box plots of scores across three dimensions: Actionability (mean: 0.6895), Criticality (mean: 0.7033), and Justice (mean: 0.7312). The right panel displays a heatmap of gap statistics between dimension pairs, with the Actionability-Justice gap (0.0704) being the most prominent, followed by the Actionability-Criticality gap (0.0608), while the Criticality-Justice gap (0.0514) shows the best balance. Darker colors in the heatmap indicate larger dimensional gaps.
}
\label{fig2}
\end{figure*}

\begin{figure*}[ht]
\centering
\includegraphics[width=\textwidth]{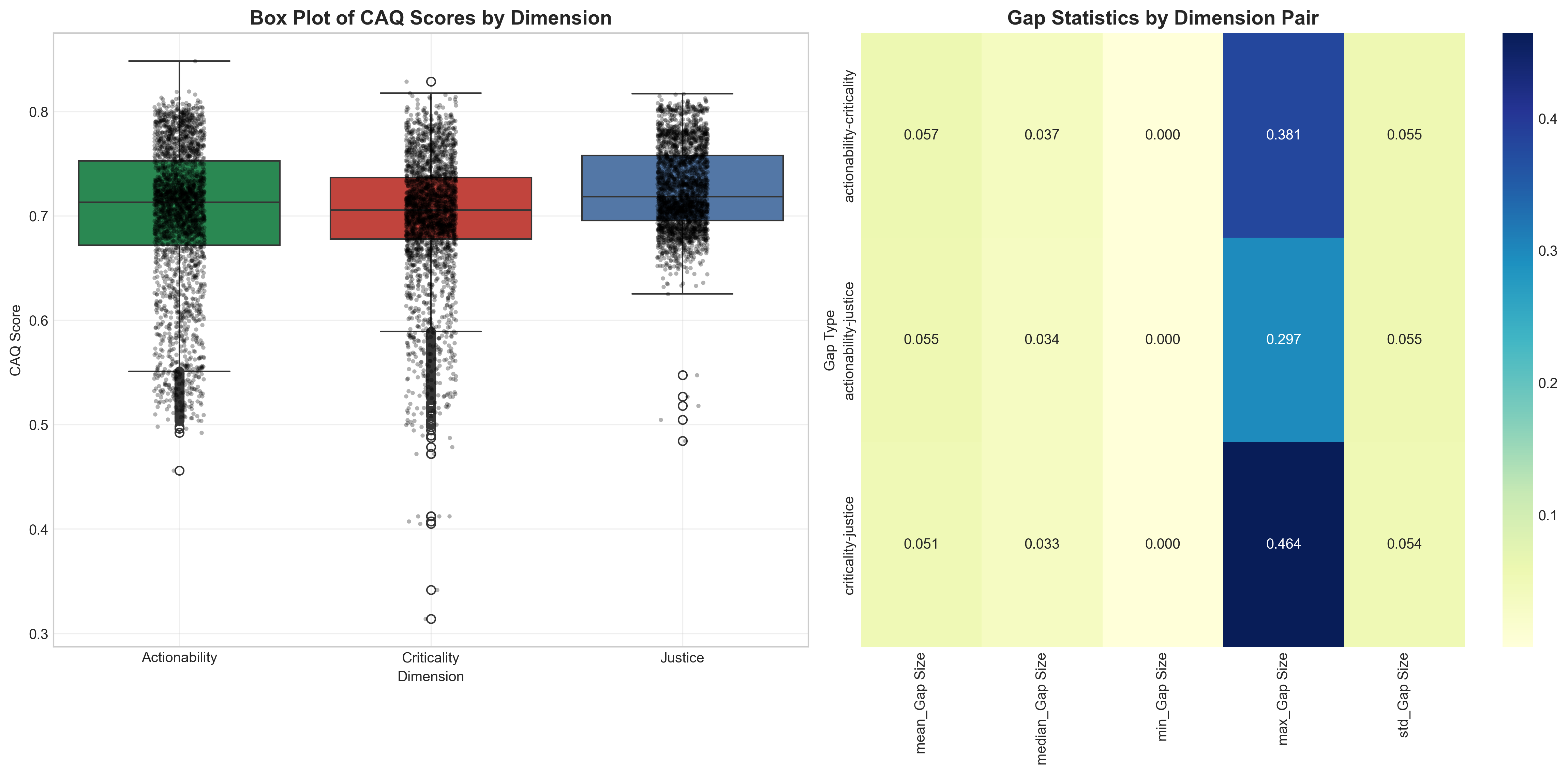}
\caption{Gap Analysis for \texttt{Qwen QwQ 32B}'s CAQ Score Performance. The left panel shows box plots of scores across three dimensions: Actionability (mean: 0.7003), Criticality (mean: 0.6990), and Justice (mean: 0.7257). The right panel displays a heatmap of gap statistics between dimension pairs, with the Actionability-Criticality gap (0.0572) being the most significant, followed by the Actionability-Justice gap (0.0548), while the Criticality-Justice gap (0.0510) shows the best balance. The analysis reveals a more pronounced variance in dimensional scores compared to other models, with a high number of messages falling into the large gap category across all dimension pairs.}
\label{fig3}
\end{figure*}

\begin{figure*}[ht]
\centering
\includegraphics[width=\textwidth]{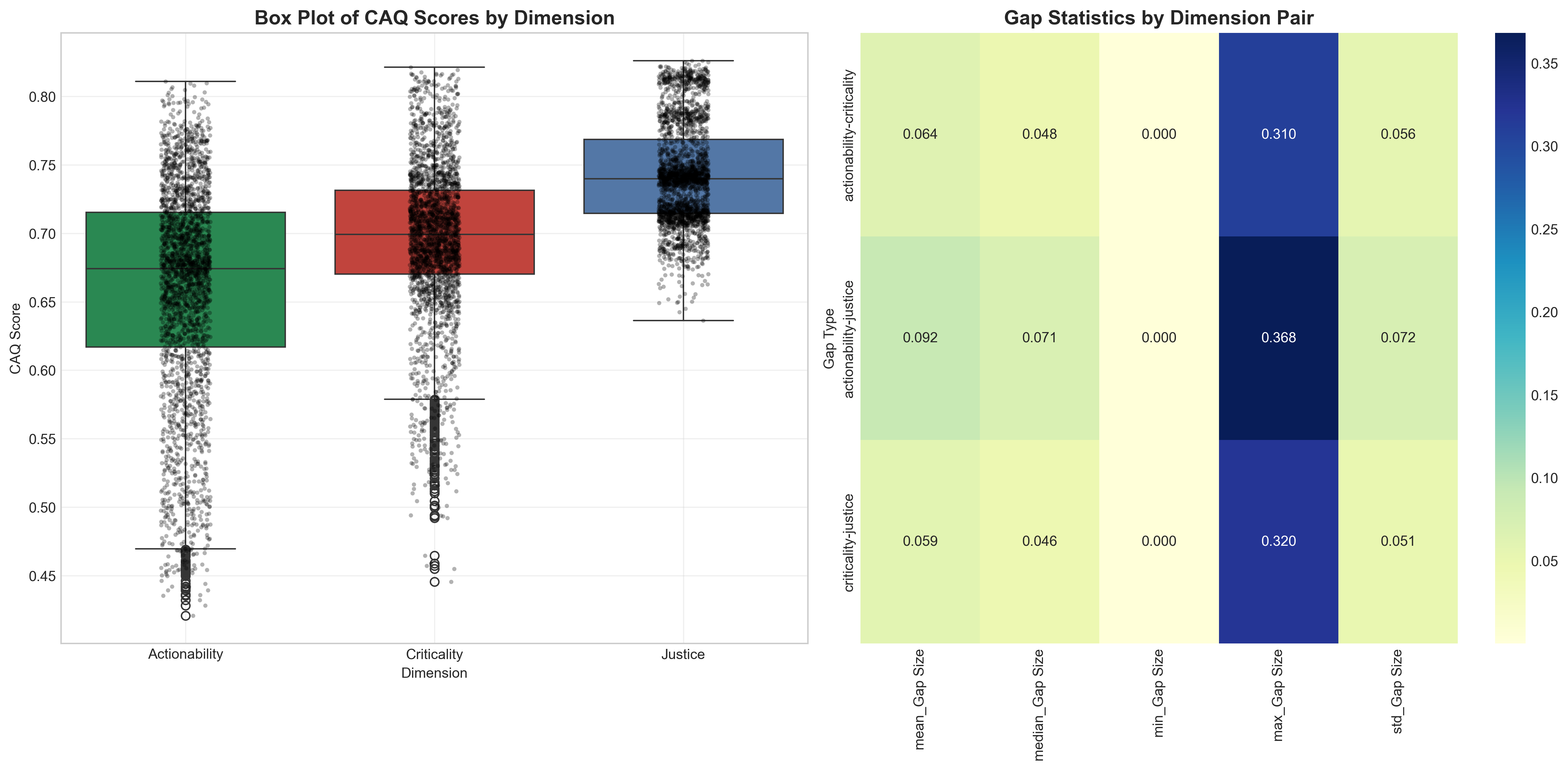}
\caption{Gap Analysis for \texttt{LlaMA 3.3 70B}'s CAQ Score Performance. The left panel shows box plots of scores across three dimensions: Actionability (mean: 0.6591), Criticality (mean: 0.6958), and Justice (mean: 0.7432). The right panel displays a heatmap of gap statistics between dimension pairs, with the Actionability-Justice gap (0.0918) being the most substantial, followed by the Actionability-Criticality gap (0.0639), while the Criticality-Justice gap (0.0590) shows the best balance. The analysis reveals a significant discrepancy between Actionability and Justice dimensions, with 520 posts (20\% of the dataset) falling into the large gap category for this dimension pair.}
\label{fig4}
\end{figure*}

\begin{figure*}[ht]
\centering
\includegraphics[width=\textwidth]{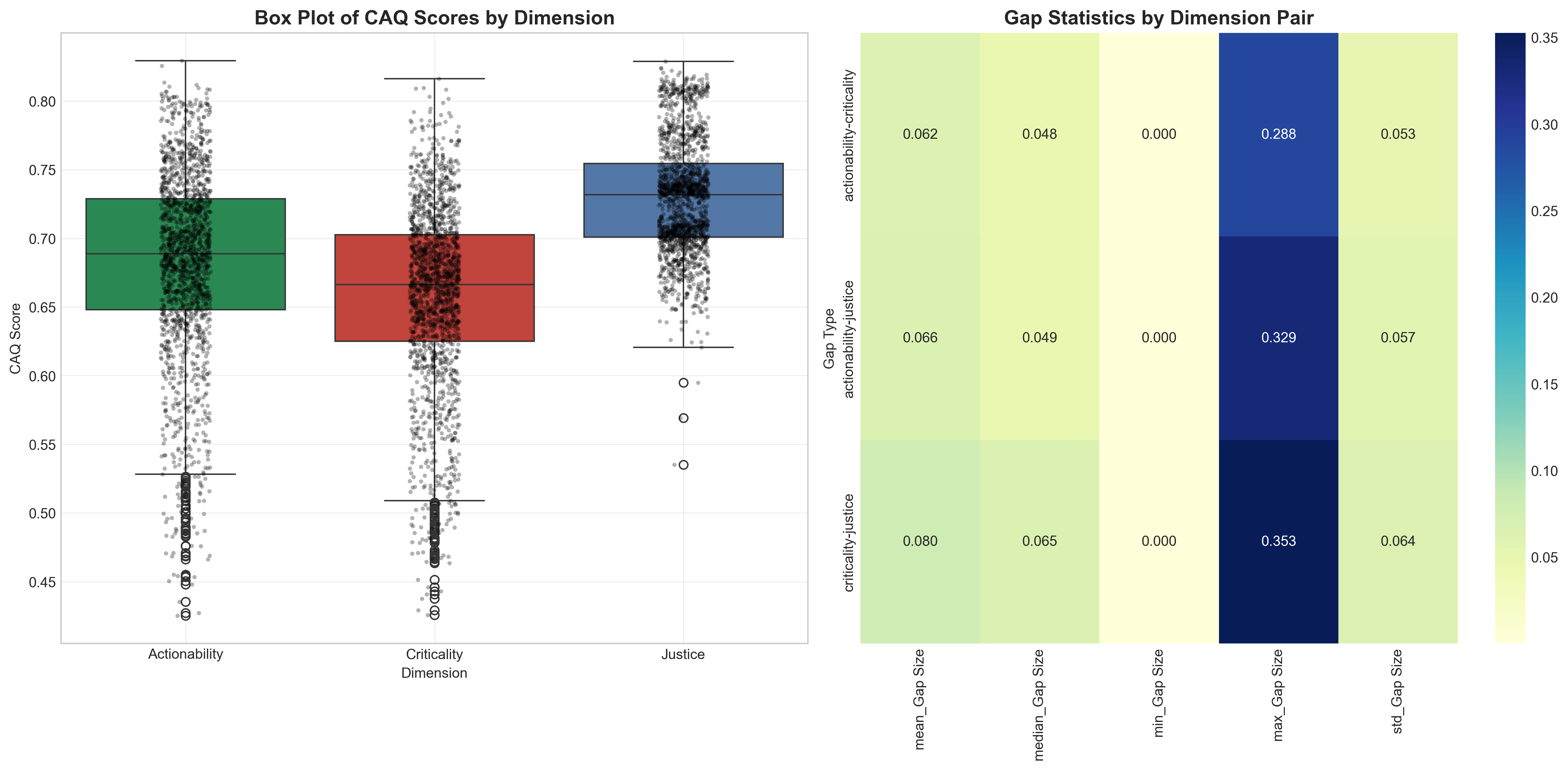}
\caption{Gap Analysis for \texttt{Gemini 2.0 Flash}'s CAQ Score Performance. The left panel shows box plots of scores across three dimensions: Actionability (mean: 0.6823), Criticality (mean: 0.6578), and Justice (mean: 0.7303). The right panel displays a heatmap of gap statistics between dimension pairs, with the Criticality-Justice gap (0.0803) being the most pronounced, followed by the Actionability-Justice gap (0.0655), while the Actionability-Criticality gap (0.0622) shows the best balance. The analysis reveals a significant discrepancy between Justice and Criticality dimensions, with 247 posts falling into the large gap category for this dimension pair, indicating a potential area for improvement in balanced climate communication.}
\label{fig5}
\end{figure*}

Figures~\ref{fig1}, \ref{fig2}, \ref{fig3}, \ref{fig4}, and \ref{fig5} collectively present a gap analysis of the CAQ scores. These figures visualize the pairwise differences between the \textit{Actionability}, \textit{Criticality}, and \textit{Justice} dimensions using box plots and heatmaps across five of our experimented models.
\label{moregap}
\subsection{Distribution of CAQ Scores for other models}
\label{3d}
We visualize the distribution of language model responses across the three CAQ dimensions, using 3D scatter plots. Table~\ref{tab:four_figures} presents these distributions for four state-of-the-art language models: GPT-4o, LLaMA 3.3 70B, Qwen QwQ 32B, and Gemini 2.0 Flash. Each point represents a single response, with its position determined by the three CAQ dimensions and color intensity corresponding to Actionability scores. This visualization allows us to identify patterns in how these models balance critical discourse, justice orientation, and actionable content when discussing climate change issues.

\begin{table*}[ht]
    \centering
    \begin{tabular}{cc}
        \includegraphics[width=0.48\textwidth]{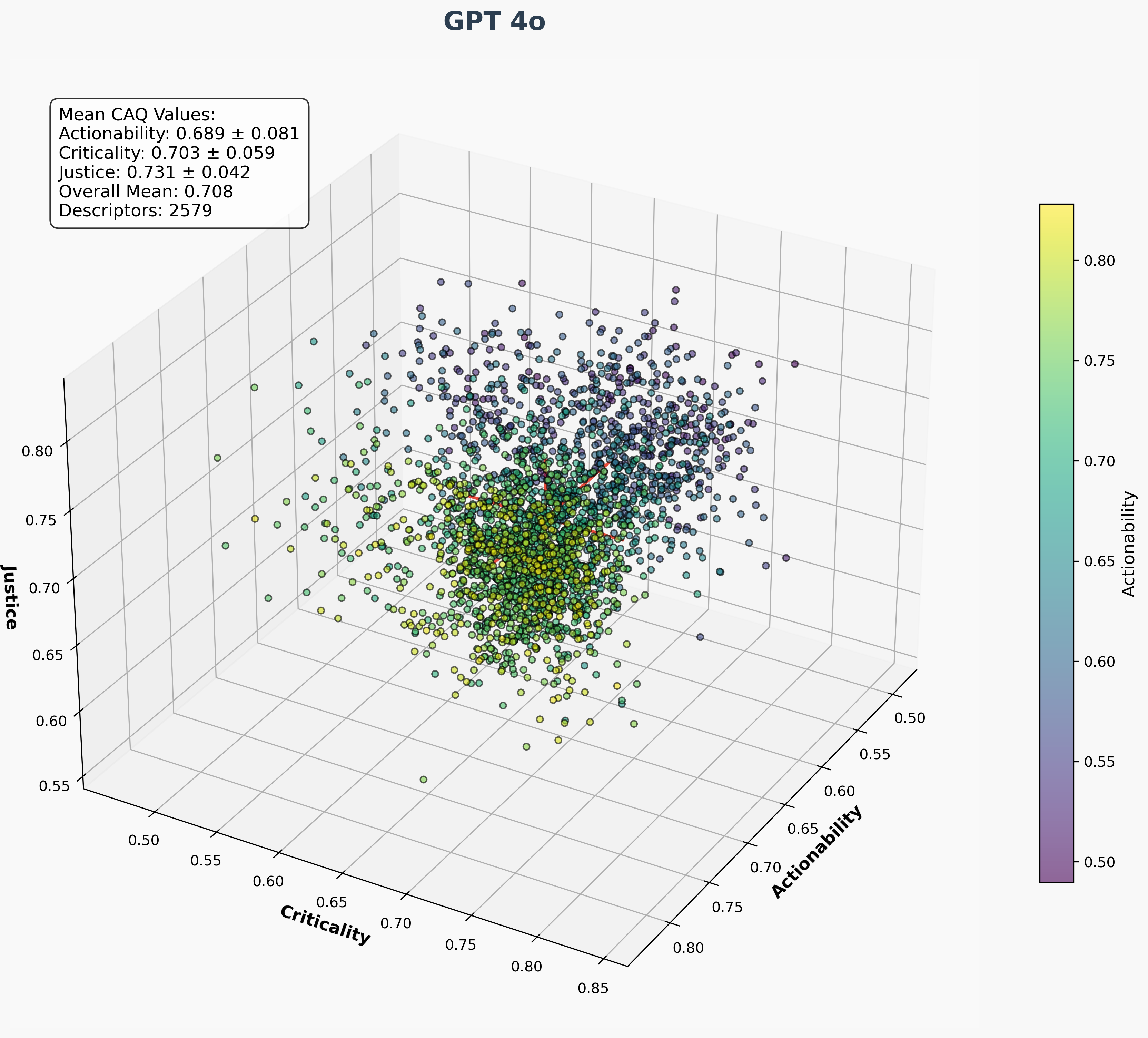} &
        \includegraphics[width=0.48\textwidth]{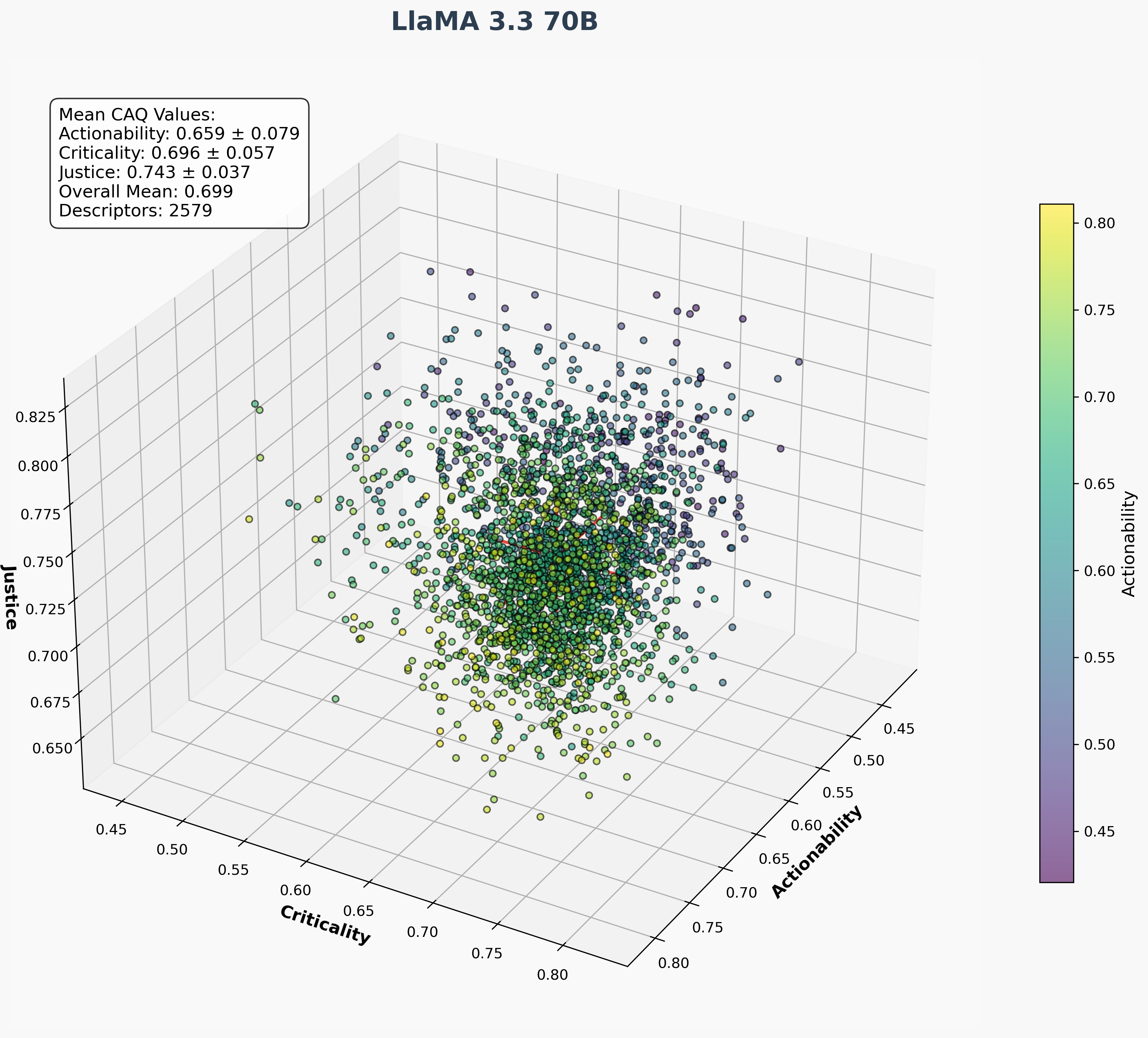} \\[10pt]
        \includegraphics[width=0.48\textwidth]{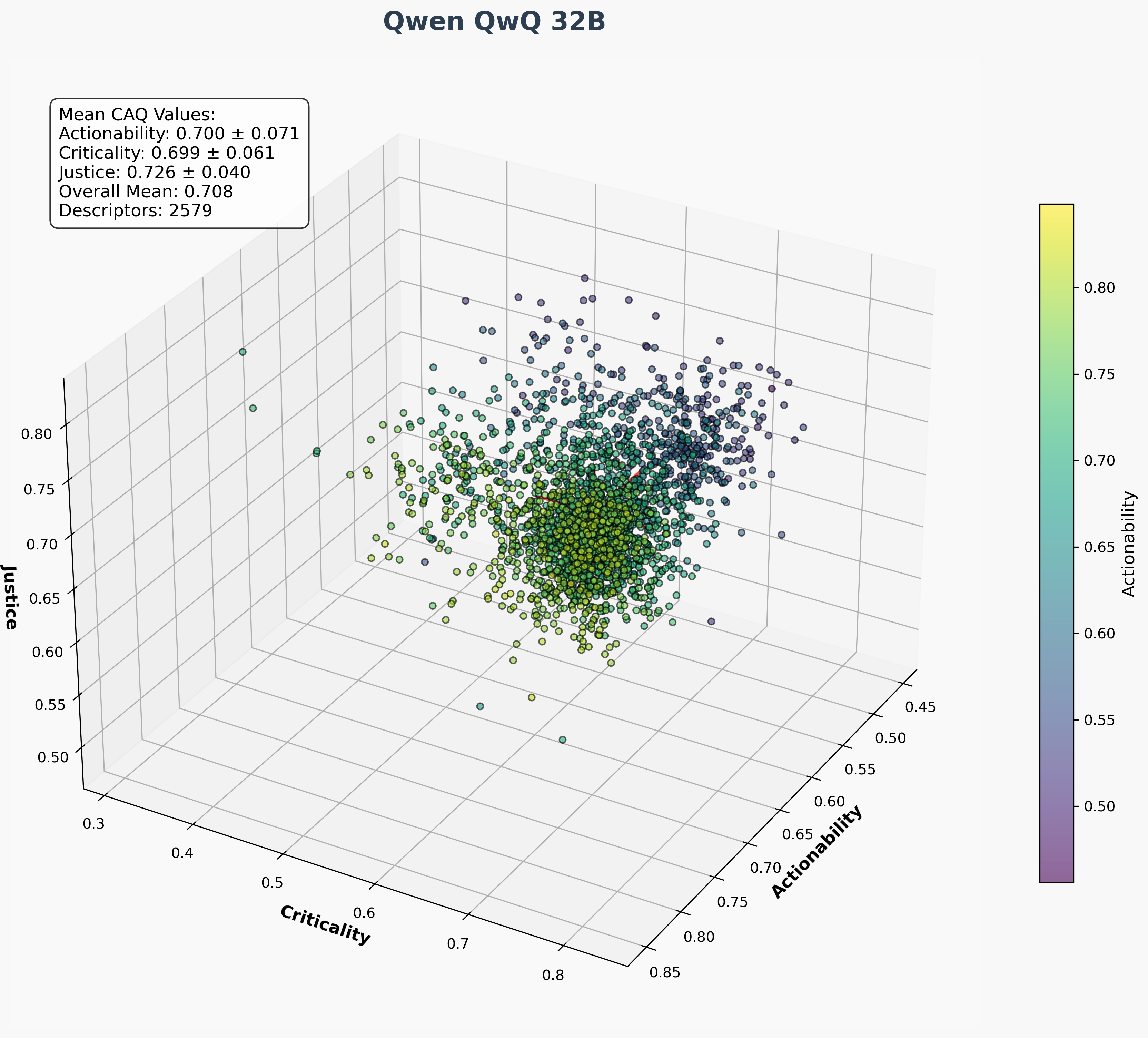} &
        \includegraphics[width=0.48\textwidth]{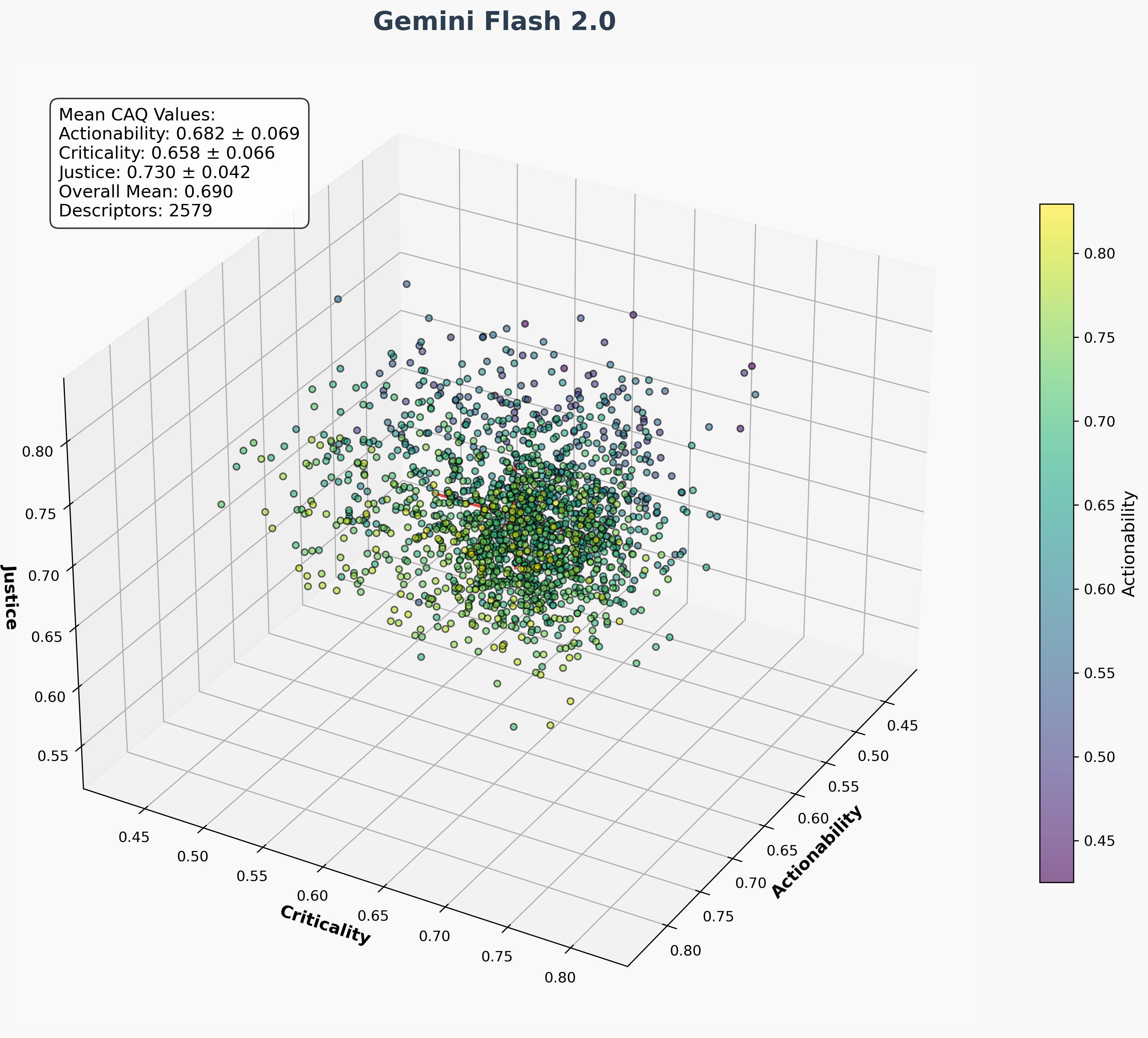} \\
    \end{tabular}
    \caption{
    3D scatter plots of CAQ scores for (top-left) GPT-4o, (top-right) LLaMA 3.3 70B, (bottom-left) Qwen QwQ 32B, and (bottom-right) Gemini 2.0 Flash. 
    The x-, y-, and z-axes represent the CAQ dimensions of \textit{Criticality}, \textit{Justice}, and \textit{Actionability}, respectively, while the color scale on the right encodes the \textit{Actionability} CAQ value, with lighter colors indicating higher values.
    Points near the center in each plot indicate relatively balanced discourse across all three dimensions, whereas outliers suggest potential gaps or imbalances. 
    We observe that GPT-4o and LLaMA 3.3 70B exhibit more points clustered near the center, indicating a higher consistency of balanced messages, while Qwen QwQ 32B and Gemini 2.0 Flash show a wider spread, reflecting greater variability in how they handle critical and justice-oriented aspects of climate communication. 
    }
    \label{tab:four_figures}
\end{table*}

\end{document}